\documentclass[conference]{IEEEtran}
\IEEEoverridecommandlockouts
\usepackage{cite}
\usepackage{amsmath,amssymb,amsfonts}
\usepackage{algorithmic}
\usepackage{graphicx}
\usepackage{textcomp}
\usepackage{xcolor}
\usepackage{amsthm}
\usepackage{algorithm}
\usepackage{multirow}
\usepackage{enumitem}
\usepackage{booktabs}
\newtheorem{myDef}{Definition}
\DeclareRobustCommand*{\IEEEauthorrefmark}[1]{%
\raisebox{0pt}[0pt][0pt]{\textsuperscript{\footnotesize\ensuremath{#1}}}}
\begin{document}

\title{A Cross-City Federated Transfer Learning Framework: A Case Study on Urban Region Profiling}

\author{
\IEEEauthorblockN{
Gaode Chen\IEEEauthorrefmark{1,2,3,4},
Yijun Su\IEEEauthorrefmark{3,4},
Xinghua Zhang\IEEEauthorrefmark{1,2}, 
Anmin Hu\IEEEauthorrefmark{3,4}, 
Guochun Chen\IEEEauthorrefmark{3,4}, 
Siyuan Feng\IEEEauthorrefmark{5}, 
\\Ji Xiang\IEEEauthorrefmark{1,2},
Junbo Zhang\IEEEauthorrefmark{3,4* \thanks{*Junbo Zhang is the corresponding author.}}, 
Yu Zheng\IEEEauthorrefmark{3,4}}
\IEEEauthorblockA{\IEEEauthorrefmark{1}Institute of Information Engineering, Chinese Academy of Sciences, Beijing, China}
\IEEEauthorblockA{\IEEEauthorrefmark{2}School of Cyber Security, University of Chinese Academy of Sciences, Beijing, China}
\IEEEauthorblockA{\IEEEauthorrefmark{3}JD iCity, JD Technology, Beijing, China \IEEEauthorrefmark{4}JD Intelligent Cities Research, Beijing, China}
\IEEEauthorblockA{\IEEEauthorrefmark{5}Department of Civil and Environmental Engineering, Hong Kong University of Science and Technology}
\IEEEauthorblockA{\{chengaode, zhangxinghua, xiangji\}@iie.ac.cn; \{suyijun.ucas, amhutx, guochun.chen3\}@gmail.com}
\IEEEauthorblockA{sfengag@connect.ust.hk; \{msjunbozhang, msyuzheng\}@outlook.com}
}

\maketitle

\begin{abstract}
Data insufficiency problems (i.e., data missing and label scarcity) caused by inadequate services and infrastructures or imbalanced development levels of cities have seriously affected the urban computing tasks in real scenarios. Prior transfer learning methods inspire an elegant solution to the data insufficiency, but are only concerned with one kind of insufficiency issue and fail to give consideration to both sides. In addition, most previous cross-city transfer methods overlook inter-city data privacy which is a public concern in practical applications. To address the above challenging problems, we propose a novel Cross-city Federated Transfer Learning framework (\textbf{CcFTL}) to cope with the data insufficiency and privacy problems. Concretely, CcFTL transfers the relational knowledge from multiple rich-data source cities to the target city. Besides, the model parameters specific to the target task are firstly trained on the source data and then fine-tuned to the target city by parameter transfer. With our adaptation of federated training and homomorphic encryption settings, CcFTL can effectively deal with the data privacy problem among cities. We take the urban region profiling as an application of smart cities and evaluate the proposed method with a real-world study. The experiments demonstrate the notable superiority of our framework over several competitive state-of-the-art methods. 
\end{abstract}


\section{Introduction}
\begin{figure}[t]
\centering
\includegraphics[width=\linewidth]{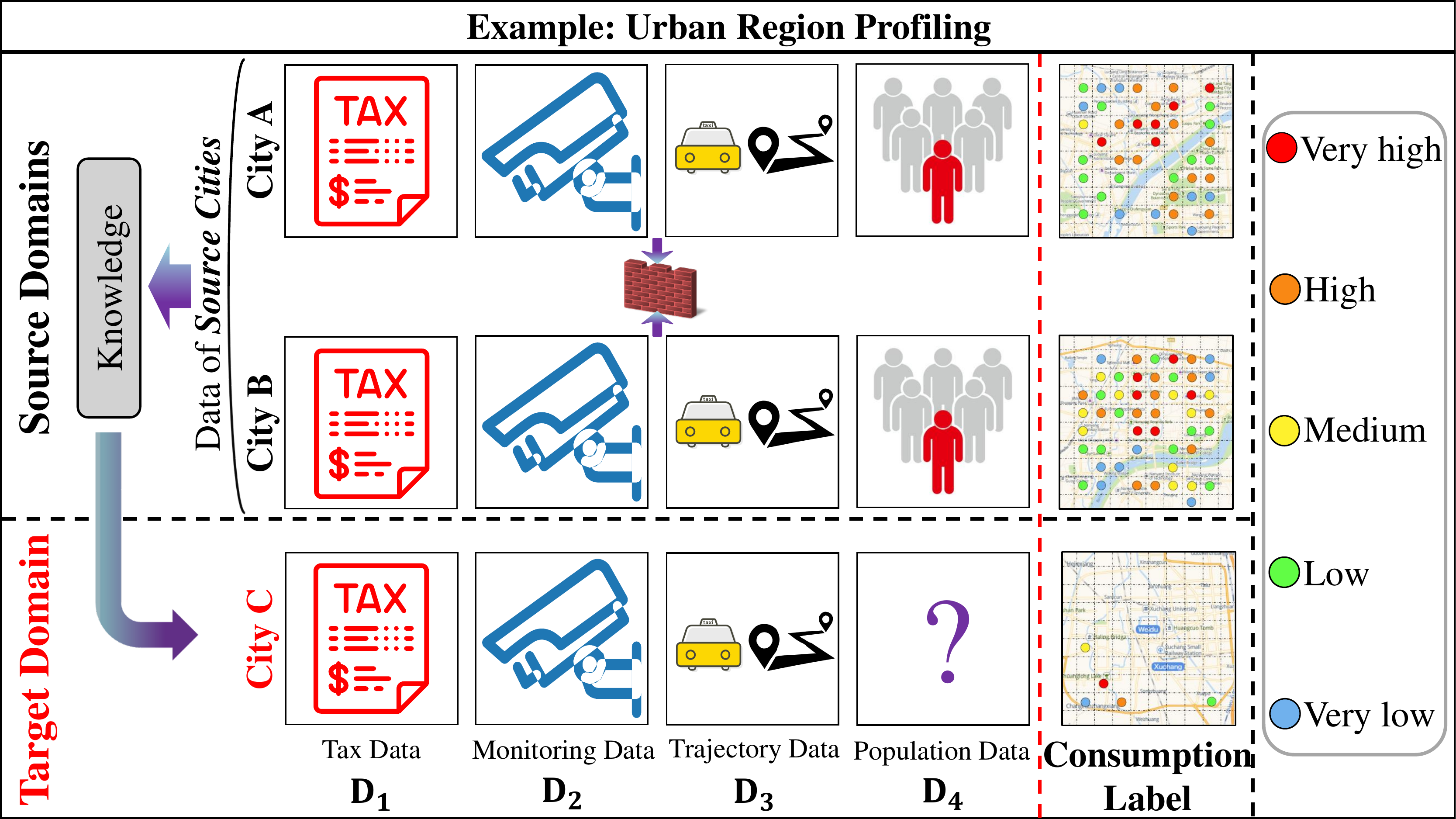}
\caption{A motivating example of our proposed framework.} 
\label{Figure 1}
\end{figure}

With the development of big data and artificial intelligence techniques, urban computing~\cite{zheng2014urban} gradually becomes the spotlight of research to bear on such urban challenges as traffic congestion and air pollution.  
Urban computing which aims at constructing a smart city, has significantly improved urban management and services.  
One of the most fundamental tasks in urban computing is accurate spatial-temporal prediction~\cite{zhang2017deep}.  
Given the complex setting of a city, we usually need to harness multi-source data to solve spatial-temporal prediction tasks.  
For example, urban region profiling aims to yield region profile vectors in the city, which describes the state of the city, such as consumption power, business popularity, and traffic convenience.  
Accurate urban region profiling needs to fully integrate and mine urban multi-source data, such as trajectory data and population density data. 
Thus, unlocking the powerful knowledge from the data collected in urban spaces is the most critical problem in urban computing. 

The crucial challenge in urban computing stems from the insufficient task data, which is due to the fact that infrastructures and services in the target city are not ready, or the city development levels are quite imbalanced.  
In the real scenario, e.g., predicting the consumption power in urban region profiling of \textit{City C} (bottom of Figure~\ref{Figure 1}), the data insufficiency is mainly embodied in two aspects: (1) \textit{Data missing problem}: a certain type of data is existing in data-rich \textit{City A} and \textit{City B}, but missing in \textit{City C}, e.g., the population density data (D4) in Figure 1, which characterizes the crowd distribution at the regional level. 
(2) \textit{Label scarcity problem}: only a few regions in \textit{City C} have ground truth labels corresponding to the task, i.e., consumption power data.  
Therefore, there is not intact and enough data to learn reliable models which brings a great obstacle for urban computing. 

Recently, transfer learning across cities provides a good perspective on the data insufficiency problems, which transfers the knowledge learned from data-rich cities (source cities) to the data-insufficient city (target city), alleviating the target data insufficiency~\cite{10.5555, yao2019learning, liu2021metastore, wu2021deep}.  
However, most prior studies only focus on a single data insufficiency problem alone or simplify the learning problems with a strong assumption. 
For example, \cite{10.5555} assumes that the target data is sparse instead of missing and then develops a cross-city transfer learning method \textit{RegionTrans} for deep spatio-temporal prediction tasks. 
\cite{wu2021deep} proposes a cross-city transfer learning framework for citywide mobile traffic prediction, which merely solves the label scarcity problem. 
Though achieving great progress, the idealized experimental settings in prior studies discount the availability when facing two kinds of data insufficiency problems simultaneously in real scenarios. 
Consequently, a credible urban computing framework should be capable of going further and addressing both {\it data missing} and {\it label scarcity} issues. 
This paper mainly focuses on exploring a cross-city transfer learning framework for disposing of these two kinds of data insufficiency problems existing in the real urban computing task. 

However, another challenge we face is that the existing cross-city transfer learning methods mentioned above ignore the data privacy protection issue. 
These methods usually involve the interaction of data among cities. 
For instance, \textit{RegionTrans}~\cite{10.5555} needs to calculate the similarity between regional data of two cities for matching. 
In recent years, the data privacy of cities becomes a public concern. 
Meanwhile, motivated by the growing interest in building smart cities, which usually need to exploit multi-source data among cities, this presents an extensive collection of open problems and challenges. 
For example, as shown in Figure~\ref{Figure 1}, we want to learn knowledge from multiple source cities (i.e., \textit{City A} and \textit{City B}) for robustness and stability, and transfer it to the target city (\textit{City C}), to predict the consumption power of regions. 
We need to analyze tax data, monitoring data, trajectory data, and population density data for more accurate results. 
As we all know, these data owned by cities are very privacy-sensitive, it is almost impossible to directly share the privacy-sensitive data across cities in a centralized way due to many legal and practical constraints. 
Thus, a framework for cross-city modeling while preserving the privacy of urban data has displayed important research and application value.  

To tackle the aforementioned challenges, we propose a novel \textbf{C}ross-\textbf{c}ity \textbf{F}ederated \textbf{T}ransfer \textbf{L}earning framework, named \textbf{CcFTL}. 
CcFTL aims to learn knowledge from multiple data-rich source cities, and transfer it to the data-insufficient target city, while preserving privacy.  
Specifically, we jointly train two modules via CcFTL for transferring knowledge to the target city: \textit{Domain Adaptive Relational-Knowledge Learning} (DARKL) module and \textit{Urban Task-specific Predicting} (UTP) module. 
The data distribution may vary considerably from city to city and even from time to time. 
Transferring data from the source cities directly to the target city can easily lead to ``negative transfer'', and it is also difficult to fit the requirement for the privacy-preserving. 
We argue that the relational knowledge among urban multi-source data is a kind of sufficiently flexible and generic knowledge across cities (detailed analysis in section \uppercase\expandafter{\romannumeral 3}.B). 
Our DARKL module learns relational knowledge by predicting ``missing data" in a regression task based on other multi-source data.  
In addition, we equip the DARKL module with a domain classifier to effectively alleviate the impact of domain discrepancies. 
Our UTP module is a classification model trained on the source data and then fine-tuned in the target city by parameter transfer, which is specific to our urban task. 

Furthermore, CcFTL treats each source city as a client to ensure the privacy of urban data during the training process of the above two modules. 
Each client trains the model based on the local urban data. 
With the elaborate adaptation of federated learning and homomorphic encryption~\cite{rivest1978data}, a central server aggregates model parameters from all clients to build a well-generalized global model for DARKL module and UTP module simultaneously. 
Afterwards CcFTL transfers this tailor-trained global model to the target city for predicting missing data and initializing urban task-specific parameters respectively. 

The contributions of this paper are summarized as follows: 
\begin{itemize}[leftmargin=*]
\item To the best of our knowledge, \textbf{CcFTL} is the first cross-city federated transfer learning framework, which takes both two kinds of data insufficiency problems and privacy protection across cities into account. 
\item With our tailor-designed \textit{Domain Adaptive Relational-Knowledge Learning} module, CcFTL is capable of obtaining the generic relational knowledge among multi-source cities for enriching the target data. And then \textit{Urban Task-specific Predicting} module is developed to transfer task parameters for alleviating the target label scarcity. 
\item We take urban region profiling as a showcase, and empirically demonstrate the notable efficacy of our proposed CcFTL on four cities from China. Moreover, a real-world case study is conducted, which provides insights into applications such as urban planning and business location recommendation. 
\end{itemize}

\section{Preliminaries}
\subsection{Concept Definition}
\begin{myDef}
  \textbf{(Region)} 
  Following previous works~\cite{zhang2016dnn, zhang2017deep}, a city $C$ is partitioned into $W_{C} \times H_{C}$ equal-size grids (e.g., 1.5km $\times$ 1.5km). 
  We treat each grid as a region $r$. 
\end{myDef}

\begin{myDef}
  \textbf{(Urban Region Profiling)} 
  Urban region profiling is similar to the concept of user profiling in Recommendation System~\cite{DBLP:conf/wsdm/GuDWY20}. 
  It abstracts the specific information of a region into labels, such as consumption power, traffic convenience, and business popularity. 
  So we treat it as a classification task, which yields region profile vectors on different grids of the city. 
  We take consumption power as an example. 
\end{myDef}

\begin{myDef}
  \textbf{(Spatial Context Feature)} 
  Spatial context feature $\mathbf{x}$ is a vector associated with a region $r$. 
  The feature are extracted from the urban multi-source data, such as road network, POIs (Points Of Interest), and population density. 
\end{myDef}

\begin{myDef}
  \textbf{(Domain)} 
  A domain~\cite{pan2009survey} consists of two components: a feature space $\mathcal{X}$ and a marginal probability distribution $P(X)$, and $X = {\mathbf{x}_1, ..., \mathbf{x}_n}$, $\mathbf{x}_i \in \mathcal{X}$. 
  In our work, a city is associated with a domain. 
  $\mathcal{X}$ is the spatial context feature space, and $P(X)$ is the spatial context feature distribution of the region in the city $X$. 
\end{myDef}

\begin{myDef}
  \textbf{(Relational Knowledge)} 
  Relational knowledge refers that some relationship among the multi-source data in the city, which is generalizable in both source and target domains. 
  We aim to learn a function $f_{RK}$ to represent this relational knowledge, which can predict the missing data based on other data.
\end{myDef} 

\subsection{Problem Definition}
Given a set of source cities $C_s$ = $\{$$c_1$, ..., $c_S$$\}$ with rich urban data and sufficient labels, and a target city $c_t$ with missing data and a very few labels. 
In other words, data insufficiency in the target city is embodied in the missing of a certain type of data and the fact that only a few regions have labels. 
Urban multi-source data cannot be shared across cities due to data sensitivity and privacy concerns. 
This work aims to utilize the data of multiple source cities, transfer the knowledge learned from these data to the target city, help the target city $c_t$ learn the prediction model $f$ based on insufficient data, and predict labels of the urban computing task. 

\textit{Consumption Power Prediction (Example).} 
We use consumption power prediction in urban region profiling as an example to illustrate the above problem concretely. 
In source cities, each region has rich data and corresponding consumption power label, including various POIs, road network, and population density data. 
In the target city, a region-level population density data is missing and only a few regions have consumption power labels. 
Later we will show that CcFTL can predict labels of all unlabelled regions in the target city. 

\section{Methodology} 
CcFTL is a cross-city federated transfer learning framework, which aims to learn knowledge from multiple data-rich source cities and transfer it to the data-insufficient target city to facilitate the task of urban computing without compromising privacy security. 
Our method has two stages. 
\textbf{\textit{Stage \uppercase\expandafter{\romannumeral 1}: federated training on source cities}}, CcFTL trains two modules based on the rich data from multiple source cities: \textit{Domain Adaptive Relational-Knowledge Learning} module and \textit{Urban Task-specific Predicting} module, which are used to learn relational knowledge among urban multi-source data and parameters specific to the urban task, respectively. 
The entire training process is based on federated learning and homomorphic encryption to protect privacy security among cities. 
\textbf{\textit{Stage \uppercase\expandafter{\romannumeral 2}: knowledge transferring on the target city}}, CcFTL transfers the learned relational knowledge and urban task-specific parameters to the target city and quickly adapts to the target domain by fine-tuning. 
We demonstrate the entire process of the CcFTL framework by taking the prediction of the consumption power in the urban region profiling as an example. 
Figure~\ref{Figure 3} gives an overview of the CcFTL framework.  

\subsection{Spatial Context Feature Extraction}
No matter which stage, we first need to process multi-source data of the source city or target city. 
The urban region profiling is highly related to the spatial context features. 
We show the process of extracting spatial features from multi-source data, including POIs, road network, and population density features. 

\textbf{Points of interest (POIs)} represent the functions and properties of regions. 
For example, a residential region may contain many communities. 
On the contrary, an office region will contain many office buildings. 
Particularly, the quantities and categories of POIs reflect the hotness of a region. 
Therefore, we extract the following POI features: (1) the number of POIs in each category $d^{pf}$, (2) the total number of POIs $d^{pn}$, and (3) the POI entropy $d^{pe}$, which can reflect the functional diversity of a region. 
It is computed as follows: 
\begin{equation}
  d^{pe} = - \sum_i \frac{d_i^{pf}}{d^{pn}} \times \log \frac{d_i^{pf}}{d^{pn}}\label{eq1}
\end{equation}
where $i$ is a POI category. 
We want to use these features to capture the hotness and functions of regions. 

\textbf{Road network} is an indicator of traffic convenience. 
We believe different roads have different contributions to urban region profiling. 
For example, the highway only lets the drivers go through the region quickly, while the living street can bring customers to shops. 
For road networks data, we consider the number of roads in each category $d^{rf}$.

\textbf{Population density} data indicates the crowd distributions in a region. 
For example, a region with many houses may have a high residential population density, while a region that includes entertainment venues may have a high consumption population density. 
We consider three kinds of region-level population density: (1) the working population density $d^{wp}$, (2) the residential population density $d^{rp}$, and (3) the consumption population density $d^{cp}$. 

In our assumption, the target city has missing the consumption population density data $d^{cp}$. 

\begin{figure}[t]
\centering
\includegraphics[width=1.05\linewidth, trim=0 0 0 90, clip]{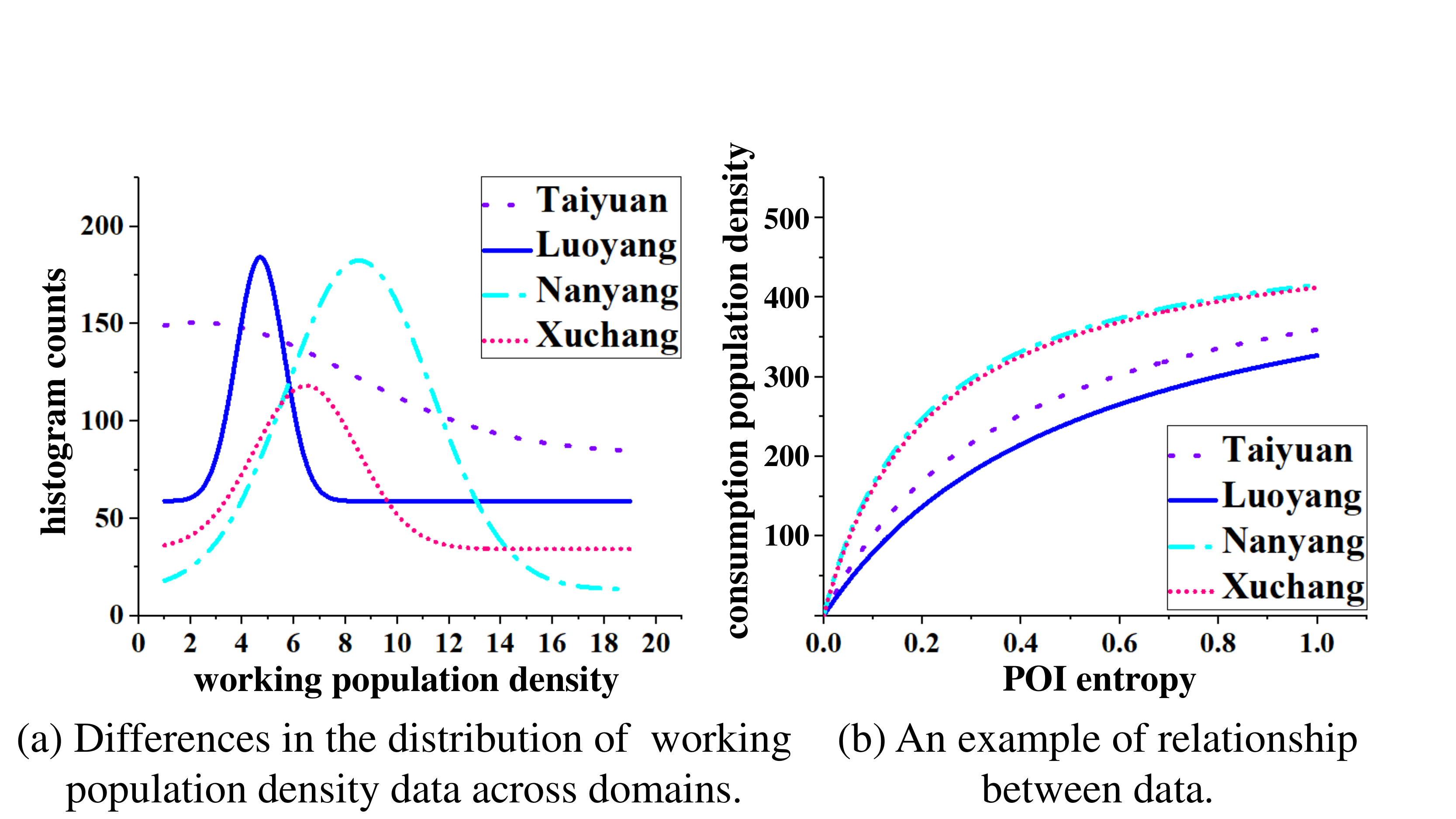}
\caption{Differences in data distribution across domains and relationships between data.}
\label{Figure 2}
\end{figure}

\begin{figure*}[t]
\centering
\includegraphics[width=\linewidth]{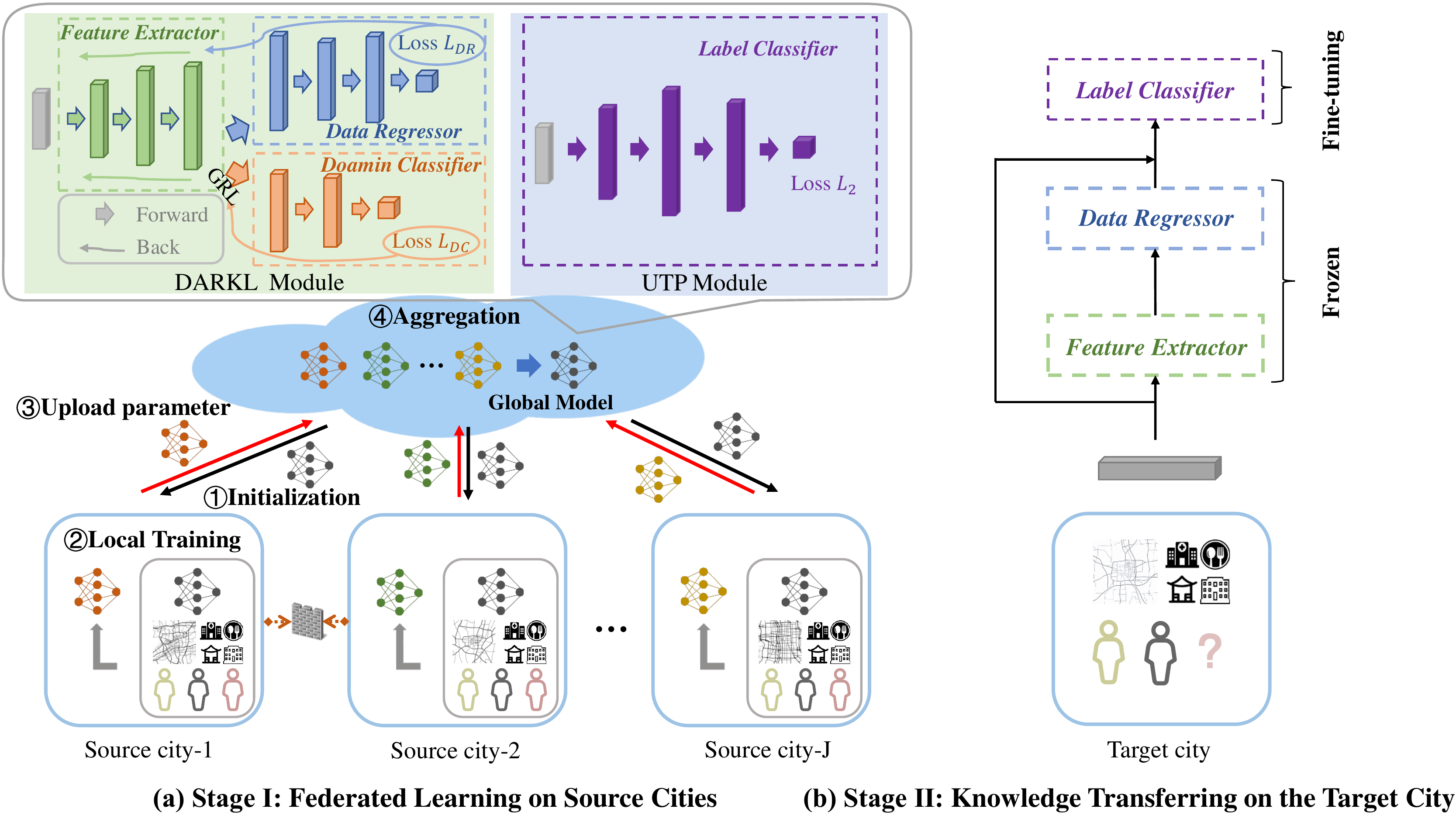}
\caption{The overview of our proposed CcFTL, which has two stages: \textit{Federated Training on Source Cities} and \textit{Knowledge Transferring on the Target City}.}
\label{Figure 3}
\end{figure*}

\subsection{Stage \uppercase\expandafter{\romannumeral 1}: Federated Training on Source Cities}
\subsubsection{Domain Adaptive Relational-Knowledge Learning (DARKL) Module}
Transferring knowledge across cities is a challenging task as the probability distributions of the data are different, i.e., $P(X_S) \neq P(X_T)$. 
As shown in Figure~\ref{Figure 2} (a), we randomly select a working population density data and show its distributions on four cities in China. 
Specifically, Figure~\ref{Figure 2} (a) demonstrates the statistics of the number of regions with different working population densities in four real cities, and the obvious discrepancy can be seen.
Thus, directly applying federated learning to cross-city scenarios without any elaborate design may not only bring negative transfer but also not work on the data insufficiency problems. 
To learn knowledge from multiple source cities to address the data insufficiency problems in the target city under the premise of privacy protection, we need to do some innovative designs for the federated learning process, which is also one of the advantages of our method. 
Some knowledge is common across domains such that they may help improve performance for the target domain or task. 
Inspired by this intuition, we argue that there are relationships among the urban multi-source data, which are similar in the source and target domain. 
We show the relationship between the POI entropy data and the consumption population density data from four cities in Figure~\ref{Figure 2} (b) as an example, which proves the similarity of the relationship across cities. 
POI entropy denotes the functional diversity of the region, and abundant POI resources will attract a large consumption population. 
Relationships between other urban data also show similarities across cities. 
Thus, relational knowledge is flexible and general enough that it can be transferred to the target city and effectively alleviates the data insufficiency problems. 

Our DARKL module can learn the relational knowledge between the missing data and other data, the input of this module is the vector $\mathbf{X}'$ $=$ $[$$d^{pf}$, $d^{pn}$, $d^{pe}$, $d^{rf}$, $d^{wp}$, $d^{rp}$$]$ containing different types of features, which is generated by the \textit{Spatial Context Feature Extraction} process, and the output of this module is the regression value for missing data. 
This module can effectively alleviate domain differences while learning relational knowledge among data, which contains three key components: \textit{Feature Extractor}, \textit{Data Regressor}, and \textit{Domain Classifier}.

\textit{Feature Extractor.} 
\textit{Feature Extractor} takes features of multiple source cities as input and outputs deep latent representations for the later regressor. 
There are two functions in Feature Extractor: (1) learning the deep latent representations of features that reduce the error of predicting missing data and (2) extracting the common representations from cities so that the \textit{Data Regressor} performs well in all cities. 
Specifically, \textit{Feature Extractor} contains $N_{fe}$ fully connected layers: 
\begin{equation}
\begin{split}
  &\mathbf{H}^{fe}_1 = \sigma\left(\mathbf{W}^{fe}_{1}\mathbf{X}' + \mathbf{b}^{fe}_{1}\right) \\
  &\mathbf{O}_{fe} = \sigma\left(\mathbf{W}^{fe}_{N_{fe}}\mathbf{H}^{fe}_{N_{fe}-1} + \mathbf{b}^{fe}_{N_{fe}}\right)
  \label{eq11}
\end{split}
\end{equation}
where the $\mathbf{W}^{fe}$ and $\mathbf{b}^{fe}$ are weight and bias of the \textit{Feature Extractor}, and $\sigma$ is the non-linear activation function for which we use $\textrm{ReLU}$. 
Finally, the high-dimensional feature representations $\mathbf{O}_{fe}$ extracted by the \textit{Feature Extractor} will be fed into the \textit{Data Regressor} and \textit{Domain Classifier} simultaneously. 

\textit{Data Regressor.} \textit{Data Regressor} is used to predict the value of missing data, which is the consumption population density in this paper. 
\textit{Data Regressor} contains $N_{dr}$ fully connected layers and a regression layer. 
The layers can be defined as: 
\begin{equation}
\begin{split}
  &\mathbf{H}^{dr}_1 = \sigma \left(\mathbf{W}^{dr}_{1}\mathbf{O}_{fe} + \mathbf{b}^{dr}_{1}\right) \\
  &\mathbf{H}^{dr}_{N_{dr}} = \sigma \left(\mathbf{W}^{dr}_{N_{dr}}\mathbf{H}^{dr}_{N_{dr}-1} + \mathbf{b}^{dr}_{N_{dr}}\right) \\
  &\hat{a} = \textrm{Sigmoid} \left(\mathbf{W}^{dr}_{N_{dr}+1}\mathbf{H}^{dr}_{N_{dr}} + \mathbf{b}^{dr}_{N_{dr}+1}\right)
  \label{eq12}
\end{split}
\end{equation}
where the $\mathbf{W}^{dr}$ and $\mathbf{b}^{dr}$ are weight and bias of the \textit{Data Regressor}, and $\sigma$ is the non-linear activation function $\textrm{ReLU}$, $\hat{a}$ is the predicted value. 

The purpose of \textit{Data Regressor} is to predict the missing data as precisely as possible. 
Thus, the parameters $\theta_{DR}$ are updated by minimizing the squared error loss $\mathcal{L}_{DR}$: 
\begin{equation}
  \mathcal{L}_{DR} = \frac{1}{N} \sum_{i = 1}^{N} {(a_i - \hat{a}_i)}^2\label{eq2}
\end{equation}
where $a_i$ is the ground truth of missing data, $N$ is the number of instances. 

\textit{Domain Classifier.} 
If we merely use \textit{Feature Extractor} and \textit{Data Regressor} to learn the relational knowledge based on multiple cities, the model may struggle with the domain shift problem. 
To address this problem, we introduce a \textit{Domain Classifier}~\cite{ganin2015unsupervised} to alleviate the feature discrepancy among different cities by using a supervised adversarial strategy. 
\textit{Domain Classifier} takes the outputs of \textit{Feature Extractor} as inputs and finally distinguishes which city the inputs come from. 
It consists of $N_{dc}$ fully connected layer and a classification layer, which can be formulated as: 
\begin{equation}
\begin{split}
  &\mathbf{H}^{dc}_1 = \sigma \left(\mathbf{W}^{dc}_{1}\mathbf{O}_{fe} + \mathbf{b}^{dc}_{1}\right) \\
  &\mathbf{H}^{dc}_{N_{dc}} = \sigma \left(\mathbf{W}^{dc}_{N_{dc}}\mathbf{H}^{dc}_{N_{dc}-1} + \mathbf{b}^{dc}_{N_{dc}}\right) \\
  &\hat{\mathbf{d}} = \textrm{Softmax}\left(\mathbf{W}^{dc}_{N_{dc}+1}\mathbf{H}^{dc}_{N_{dc}} + \mathbf{b}^{dc}_{N_{dc}+1}\right)
  \label{eq13}
\end{split}
\end{equation}
where the $\mathbf{W}^{dc}$ and $\mathbf{b}^{dc}$ are weight and bias of the \textit{Domain Classifier}, $\sigma$ denotes the non-linear activation function $\textrm{ReLU}$, $\hat{\mathbf{d}}$ is the predicted domain label. 

The parameters of \textit{Domain Classifier} $\theta_{DC}$ are optimized by minimizing the cross-entropy loss $\mathcal{L}_{DC}$:
\begin{equation}
  \mathcal{L}_{DC} = \frac{1}{N} \sum_{i = 1}^{N} \sum_{j = 1}^{|\mathcal{C}|} - \mathbf{d}_{ij} \log \hat{\mathbf{d}}_{ij}\label{eq3}
\end{equation}
where $\mathbf{\mathbf{d}}_{ij}$ is the ground truth of domain label, $|\mathcal{C}|$ is the number of domains (cities), $N$ is the number of instances. 

It is worth noting that \textit{Feature Extractor} can not handle the latent representations of the target city if the representations of cities are discriminative. 
Thus, \textit{Feature Extractor} should learn common representations rather than discriminative representations for cities' features. 
To this end, the parameters $\theta_{FE}$ of \textit{Feature Extractor} should be optimized by maximizing the loss $\mathcal{L}_{DC}$. 
To implement this, we introduce the gradient reversal layer (GRL)~\cite{ganin2015unsupervised} between \textit{Feature Extractor} and \textit{Domain Classifier}. 
During the forward propagation, there is no influence when features pass through GRL. 
But during the backward propagation, GRL will multiply the gradient from the \textit{Domain Classifier} by $-\lambda$ and pass it to \textit{Feature Extractor}. 

To summarize, we optimize the DARKL module parameters $\theta_{RK}$ by minimizing the total loss 
$\mathcal{L}_1$: 
\begin{equation}
  \mathcal{L}_{1} = \mathcal{L}_{DR} - \lambda \mathcal{L}_{DC}\label{eq4}
\end{equation}
where $\lambda$ is a hyper-parameter to control the trade-off between regressor loss $\mathcal{L}_{DR}$ and classifier loss $\mathcal{L}_{DC}$. 

\subsubsection{Urban Task-specific Predicting (UTP) Module}
This module is designed for the specific urban computing task. 
We treat it as a classification task, which can be applied to predict urban region profiling, and our work takes the consumption power prediction as an example. 
Thus, the inputs are tuples of spatial context features $\mathbf{X}$ $=$ $[$$d^{pf}$, $d^{pn}$, $d^{pe}$, $d^{rf}$, $d^{wp}$, $d^{rp}$, $d^{cp}$$]$ for regions, the outputs are the consumption power categories. 
UTP module contains $N_{utp}$ fully connected layers and a prediction layer. We can obtain the predicted result by the following formula: 
\begin{equation}
\begin{split}
  &\mathbf{H}^{utp}_1 = \sigma \left(\mathbf{W}^{utp}_{1}\mathbf{X} + \mathbf{b}^{utp}_{1}\right) \\
  &\mathbf{H}^{utp}_{N_{utp}} = \sigma \left(\mathbf{W}^{utp}_{N_{utp}}\mathbf{H}^{utp}_{N_{utp}-1} + \mathbf{b}^{utp}_{N_{utp}}\right) \\
  &\hat{\mathbf{y}} = \textrm{Softmax}\left(\mathbf{W}^{utp}_{N_{utp}+1}\mathbf{H}^{utp}_{N_{utp}} + \mathbf{b}^{utp}_{N_{utp}+1}\right)
  \label{eq5}
\end{split}
\end{equation}
where $\mathbf{W}^{utp}$ and $\mathbf{b}^{utp}$ are the trainable parameters of the Urban Task-specific Predicting module, and $\hat{\mathbf{y}}$ is the predicted probability value of the consumption power category. 

We optimize the UTP module parameters $\theta_{task}$ by minimizing the cross-entropy loss function $\mathcal{L}_{2}$: 
\begin{equation}
  \mathcal{L}_{2} = \frac{1}{N} \sum_{i = 1}^{N} \sum_{j = 1}^{|\mathcal{K}|} - \mathbf{y}_{ij} \log \hat{\mathbf{y}}_{ij}\label{eq6}
\end{equation}
where $\mathbf{y}_{ij}$ is the ground truth of the consumption power, $|\mathcal{K}|$ is the number of consumption power categories, and $N$ is the number of instances. 

\subsubsection{Training Process}

CcFTL adopts the horizontal federated learning paradigm to achieve encrypted model training and sharing. 
Our goal is to train a well-generalized global model for the DARKL module and UTP module simultaneously based on the local spatial context data of each participating city, with the help of a central server. 

In the federated learning environment, each source city is considered a client, and each client trains the DARKL module and UTP module jointly. 
The data from each city cannot be shared and should be kept locally. 
There is a central server that aggregates the model parameters uploaded by each city in each iteration and transmits the aggregated global model to each source city for the next iteration. 
We introduce the FedAVG algorithm~\cite{mcmahan2017communication} as the core of the secure parameter aggregation mechanism to collect model parameters from different clients and improve them for our scenarios. 
Our algorithm is an iterative process, for the $i$-th round of training, the models of clients participating in the training will be updated to the new global one. 
It consists of four steps: 
\begin{itemize}[leftmargin=*]
\item[i)] \textit{Step 1, Initialization}: The central server initializes the parameters $\theta_{RK}$ and $\theta_{task}$ of the global model and distributes the copy of the global model to all clients; 
\item[ii)] \textit{Step 2, Local Training}: Each client trains spatial context data locally and updates $\theta_{RK}$ and $\theta_{task}$ for $E$ epochs with SGD optimizer;
\item[iii)] \textit{Step 3, Upload Parameter}: Each client uploads the updated model to the central server. Note that this step does not share any local data or information but the model parameters are encrypted by homomorphic encryption;
\item[iv)] \textit{Step 4, Aggregation}: The central server aggregates the updated parameters uploaded by all clients by the secure parameter aggregation mechanism to build a new global model, and then distributes the new global model to each client for the next round of updates until convergence. 
\end{itemize}

The pseudocode for training process of the CcFTL is presented in Algorithm~\ref{alg:algorithm}. 

CcFTL overcomes the limitation of data islanding among cities. 
Our DARKL module can learn general relational knowledge among urban multi-source data and effectively reduce the impact of domain differences.  
Relational knowledge can be transferred to the target city and alleviate its data insufficiency problems. 
On the other hand, our UTP module is also transferable and can be used for the initial parameters of urban computing tasks in the target city. 

\subsection{Stage \uppercase\expandafter{\romannumeral 2}: Knowledge Transferring on the Target City}
Our framework CcFTL uses transfer learning to address the challenge of data insufficiency for the target city. 
Figure~\ref{Figure 3} (b) shows the process of transfer learning in the target city. 
We train the \textit{Domain Adaptive Relational-Knowledge Learning} module and \textit{Urban Task-specific Predicting} module based on multi-source data from multiple source cities and apply them in the target city. 

\begin{algorithm}[tb]
\flushleft
\caption{The training procedure of CcFTL}
\label{alg:algorithm}
\textbf{Input}: Spatial context data $X_i$ of city $i$, initial global model parameters $\overline{\theta}_{RK}^{(0)}$ 
and $\overline{\theta}_{task}^{(0)}$, the maximum number of global rounds $R_g$, the epochs of client's local training $E$ and the learning rate $\eta$. \\
\textbf{Output}: Trained global model parameters $\overline{\theta}_{RK}^{(R_g)}$ and $\overline{\theta}_{task}^{(R_g)}$. \\
\textbf{Server executes:} 
\begin{algorithmic}[1]
\FOR {each client $i$ $\in$ $|\mathcal{C}|$ in parallel}
    \STATE Initialize client model $\theta_{RK, i}^{(0)} = \overline{\theta}_{RK}^{(0)}$, $\theta_{task, i}^{(0)} = \overline{\theta}_{task}^{(0)}$. \\
\ENDFOR
\FOR {global round $r_g = 1, 2, ..., R_g$}
    \FOR {each client $i$ $\in$ $|\mathcal{C}|$ in parallel}
    \STATE $\theta_{RK, i}, \theta_{task, i} \gets ClientUpdate(i, E, \eta)$. \\
    \ENDFOR
    \STATE $\overline{\theta}_{RK}^{(r_g)} \gets \sum_{i \in |\mathcal{C}|} \frac{N_i}{N} \theta_{RK, i}$, $\overline{\theta}_{task}^{(r_g)} \gets \sum_{i \in |\mathcal{C}|} \frac{N_i}{N} \theta_{task, i}$. \\
    \FOR {each client $i$ $\in$ $|\mathcal{C}|$ in parallel}
    \STATE Initialize client model $\theta_{RK, i}^{(0)} = \overline{\theta}_{RK}^{(r_g)}$, $\theta_{task, i}^{(0)} = \overline{\theta}_{task}^{(r_g)}$. \\
    \ENDFOR
\ENDFOR
\end{algorithmic}
\textbf{ClientUpdate($i$, $E$, $\eta$):} 
\begin{algorithmic}[1]
\FOR {client epoch $e = 1, 2, ..., E$}
    \STATE $\theta_{RK, i}^{(e)} \gets \theta_{RK, i}^{(e - 1)} - \eta \nabla_{\theta_{RK, i}^{(e - 1)}} \mathcal{L}_1$, $\theta_{task, i}^{(e)} \gets \theta_{task, i}^{(e - 1)} - \eta \nabla_{\theta_{task, i}^{(e - 1)}} \mathcal{L}_2$. \\
\ENDFOR
\STATE $\theta_{RK, i} = \theta_{RK, i}^{(E)}$, $\theta_{task, i} = \theta_{task, i}^{(E)}$. \\
\STATE \textbf{return} $\theta_{RK, i}, \theta_{task, i}$ to server. \\
\end{algorithmic}
\end{algorithm}

Firstly, due to relational knowledge is general across cities, we feed the spatial context features $\mathbf{X}_t$ $=$ $[$$d_t^{pf}$, $d_t^{pn}$, $d_t^{pe}$, $d_t^{rf}$, $d_t^{wp}$, $d_t^{rp}$$]$ of the target city into the DARKL module, where the missing data is $d_t^{cp}$. 
We can restore the missing data to some extent as follows:
\begin{equation}
  \hat{d}_t^{cp} = f \left(\mathbf{X}_t; \theta_{RK}\right)\label{eq7}
\end{equation}
where $\theta_{RK}$ are parameters of the DARKL module. 

And then, based on the spatial context data $X_t$ and predicted missing data $\hat{d}_t^{cp}$, we utilize the UTP module to predict the consumption power in urban region profiling. 
\begin{equation}
  \hat{\mathbf{y}}_t = f \left(\mathbf{X}_t, \hat{d}_t^{cp}; \theta_{task}\right)\label{eq8}
\end{equation}

During the transfer learning process, since the target city does not have the ground truth value of the missing data, the label distributions of urban computing tasks still exist differences across cities. 
Thus, we only use the trained DARKL module for dealing with the missing data and keep the parameters frozen in the target city. 
Then we fine-tune the UTP module by minimizing the following loss function based on very few labels in the target city: 
\begin{equation}
  \mathcal{L}_{t} = \frac{1}{M} \sum_{i = 1}^{M} \sum_{j = 1}^{|\mathcal{K}|} - \mathbf{y}_t \log \hat{\mathbf{y}}_{t}\label{eq9}
\end{equation}
where $\mathbf{y}_{t}$ is the ground truth of the consumption power, $\hat{\mathbf{y}}_{t}$ is the predicted label, $|\mathcal{K}|$ is the number of consumption power categories, $M$ represents the number of labeled instances. 

\subsection{Privacy Analysis}
In this section, we mainly focus on the privacy analysis of our framework CcFTL against external attackers, showing whether it can protect the privacy security of urban data during the cross-city federated transfer learning process.  

Firstly, urban data owned by cities is very privacy-sensitive, and even if they are shared through privacy-preserving technologies, there is a risk of privacy leakage. 
Our framework CcFTL ensures that urban data never come out of the local storage during the federated transfer learning process which alleviates privacy concerns. 

Secondly, the transmission of gradients and partial parameters in \textit{Stage \uppercase\expandafter{\romannumeral 1}} may lead to indirect privacy leakage, since external attackers will use inference attacks~\cite{nasr2019comprehensive, luo2021feature}. 
In addition, Homomorphic encryption refers to an encryption mechanism that encodes parameters before addition or multiplication and produces an equivalent result compare to uncode function. 
Thus, we apply homomorphic encryption in CcFTL to hinder information leakage during parameter exchange between client and server, which makes it possible to compute aggregation on the encrypted server-side. 

Finally, the model parameters transferred to the target city are learned from multiple source cities, from which it is difficult for an external attacker to infer the data distribution of any one source city, thus ensuring the security and robustness of the transfer process. 

\section{Experiments}
In this section, we evaluate the CcFTL on the task of urban region profiling. 
In experiments, CcFTL transfers knowledge from multiple source cities in China, e.g., Luoyang and Taiyuan, to improve accuracies of consumption power prediction in the target city, Xuchang, which faces the data insufficiency problems. 

\subsection{Experimental Settings}
\subsubsection{Data Description}
We collect four categories of data from four cities (i.e., Luoyang, Nanyang, Xuchang, and Taiyuan) in China, ranging from November 2020 to November 2021, as shown follow: 
\begin{itemize}[leftmargin=*]
\item \textbf{Road Networks \& POIs Data.} 
Road network data from Bing Maps\footnote{https://www.bing.com/maps.} contains road segments each of which is described with its endpoints and level of capacity. 
POIs data from Bing Maps indicates the name, address, coordinates, and category of each point of interest. 
\item \textbf{Population Density Data.} 
We collect three types of region-level population density data from governments of cities\footnote{The data is not used due to our data confidential agreement.}, including the working population density, residential population density, and consumption population density. 
We assume that the target city is missing consumption population density data. 
Details of datasets are summarized in Table~\ref{tab1}. 
\item \textbf{Consumption Power Data.} 
The consumption power data, as the labels of the task, are also collected from governments of cities. 
The consumption power values range from one to five, corresponding to five different consumption power levels, i.e., "Very high", "High", "Medium", "Low", and "Very low", respectively.  
\end{itemize}

Each city is partitioned into grid regions in the size of $152m \times 152m$. 
To evaluate our model comprehensively, we conduct experiments with four city pairs: (Nanyang, Xuchang) $\to$ Luoyang, (Taiyuan, Luoyang) $\to$ Nanyang, (Luoyang, Taiyuan) $\to$ Xuchang, and (Xuchang, Nanyang) $\to$ Taiyuan. 
In source cities, we randomly choose 80\% of regions as training data and the rest as test data. 
In the target city, we randomly select only 20\% of regions for training or fine-tuning and the rest as the test data. 

\subsubsection{Evaluation Metrics}
\begin{table}[t]
\caption{Details of Datasets.}
\resizebox{\linewidth}{!}{
\begin{tabular}{ccccc}
\toprule
\multirow{2}{*}{\textbf{Data}}    & \multicolumn{4}{c}{\textbf{City}}              \\
                  & Luoyang & Nanyang & Xuchang & Taiyuan \\ \midrule
\#of regions          & 5,466        & 5,322        & 3,169        & 3,982        \\ \midrule
\#of POIs    & 130,628        & 108,189        & 51,842        & 20,716        \\
\#of roads            & 22,434        & 17,111        & 11,815        & 33,965        \\ 
\#of working pop.     & 644,706        & 595,074        & 280,195        & 212,348        \\
\#of residential pop.  & 1,197,048        & 1,217,450        & 501,845        & 394,004        \\
\#of consumption pop.  & 303,804        & 303,883        & 175,093        & 199,662        \\ \bottomrule
\end{tabular}}
\label{tab1}
\end{table}

Our purpose is to restore missing data and predict the consumption power in the target city. 
Thus, we adopt both regression and classification metrics to demonstrate the performance of CcFTL and baseline methods: 

\textbf{MAE \& MSE.} 
MAE and MSE are the mean absolute error and mean squared error between the ground truth and predicted values, respectively. 
The lower value of MAE and MSE implies higher accuracy of a regression model. MAE and MSE are formulated as: 
\begin{align}
  MAE = \frac{1}{m} \sum_{i = 1}^m |y_i - \hat{y}_i|
\end{align}
\begin{align}
  MSE = \frac{1}{m} \sum_{i = 1}^m {(y_i - \hat{y}_i)}^2
\end{align}
where $y_i$ is the ground truth, and $\hat{y}_i$ is the predicted value.

\textbf{Precision \& Recall \& F1.} 
We use precision, recall, and F1 score to evaluate the performance of consumption power prediction by comparing the predicted $E_P$ to the ground truth $E_G$. 
Precision, Recall and F1 score are defined as: 
\begin{align}
  Precision = \frac{|E_P \cap E_G|}{|E_P|}
\end{align}
\begin{align}
  Recall = \frac{|E_P \cap E_G|}{|E_G|}
\end{align}
\begin{align}
  F_1 \ score = \frac{2 * \textit{Precision} * \textit{Recall}}{\textit{Precision} + \textit{Recall}}
\end{align}
The larger value of precision, recall, and F1 score indicates that the methods predict consumption power more accurately. 
Note that precision, recall, and F1 score used in this paper refer to macro metrics. 

\subsubsection{Baseline Algorithms} 
We compare our model with the following three categories of methods: Target Only methods, Cross-city Transfer, and Cross-city Federated Transfer baselines. 
Note that, for Target Only baselines, we only use the limited data of the target city to train the model, showing performance under data insufficiency (data missing and label scarcity) conditions.
For Cross-city Transfer baselines, we learn a well-generalized initialization of a deep model from multiple source cities and fine-tune it in the target city regardless of privacy preservation concerns.  
For Cross-city Federated Transfer baselines, considering the data privacy protection of the knowledge learning and transfer process, we directly combine federated learning algorithms with deep learning models, but no other elaborate designs. 

\noindent \textbf{Target Only baselines:}

\begin{itemize}[]
  \item \textbf{XGBoost~\cite{chen2016xgboost}}: A powerful machine learning algorithm under the Gradient Boosting framework, is one of the most popular and effective methods in the industry.
  \item \textbf{CNN~\cite{kalchbrenner2014convolutional}}: A convolutional neural network consists of convolution layers, pooling layers and fully connected layers, which extract low-order to high-order features through convolution kernels one by one. 
  \item \textbf{MLP}: A feed-forward artificial neural network model, containing input layer, hidden layer and output layer, with the high  capacity to model nonlinear relationships. 
\end{itemize}

\noindent \textbf{Cross-city Transfer baselines:}

\begin{itemize}[]
  \item \textbf{MetaST~\cite{yao2019learning}}: A competitive cross-city transferred method to address spatio-temporal prediction problems in the data-scarcity target city. 
  The method learns a well-generalized initialization of the base model via meta-learning from multiple source cities and learns a global pattern-based spatio-temporal memory for transfer to the target city. 
  \item \textbf{MetaStore~\cite{liu2021metastore}}: A state-of-the-art task-adaptative model-agnostic meta-learning framework that also learns available knowledge from multiple source cities and transfers it to data-insufficient target city to improve prediction performance on urban computing tasks. 
  This method learns a set of meta-learned initializations from multiple source cities, which is capable of tackling complex data distributions and accelerating the adaptation in target city.
\end{itemize}

Note that neither MetaST nor MetaStore take into account the issue of data privacy protection and are both trained in a centralized manner. 

\noindent \textbf{Cross-city Federated Transfer baselines:}

\begin{itemize}[]
  \item \textbf{CNN+FedAvg}: A CNN model for the specific urban computing task trained with the Federated Averaging algorithm~\cite{mcmahan2017communication} on multiple source cities. And then, we transfer it to the target city for fine-tuning to address the urban computing task. 
  \item \textbf{MLP+FedAvg}: A MLP model for the specific urban computing task trained with the FedAvg algorithm on multiple source cities. It differs from CcFTL in that it does not have some elaborate designs for cross-city federated transfer scenarios, such as relational knowledge learning under federated conditions. 
\end{itemize}

Data insufficiency in the target city is embodied in two aspects: data missing problem and label scarcity problem. 
Thus, the data imputation methods are not suitable for such scenarios, and our method cannot compare with these methods. 

Besides, there are multiple implementations of our cross-city federated transfer learning framework depending on the network structure: 

\begin{itemize}[]
  \item \textbf{CcFTL(MLP)}: A model as elaborated in Section \uppercase\expandafter{\romannumeral 3}, which uses MLP as the network structure under the CcFTL framework. 
  \item \textbf{CcFTL(CNN)}: A model with the same architecture as CcFTL, but using CNN as the network structure instead of MLP, including the DARKL module and UTP module. 
\end{itemize}

\begin{table*}[t]
\caption{Comparisons with baselines on four pairs of cities. The best performance in each column is bolded number.}
\resizebox{\linewidth}{!}{
\begin{tabular}{lcccccccccccc}
\toprule
\multicolumn{1}{c}{}                                & \multicolumn{3}{c}{\textbf{(Ny, Xc) $\to$ Ly}} & \multicolumn{3}{c}{\textbf{(Ty, Ly) $\to$ Ny}} & \multicolumn{3}{c}{\textbf{(Ly, Ty) $\to$ Xc}} & \multicolumn{3}{c}{\textbf{(Xc, Ny) $\to$ Ty}} \\ \cmidrule(r){2-4} \cmidrule(r){5-7} \cmidrule(r){8-10} \cmidrule(r){11-13}
\multicolumn{1}{c}{}                                & P   & R    & F1  & P   & R    & F1  & P   & R    & F1  & P   & R    & F1  \\ \midrule
\multicolumn{13}{l}{\textbf{Target Only baselines}}      \\
XGBoost    & 0.221	& 0.204	& 0.171	& 0.180	& 0.209	& 0.173	& 0.196	& 0.211	& 0.186	& 0.117	& 0.169	& 0.147 \\
CNN    & 0.171	& 0.206	& 0.174	& 0.201	& 0.204	& 0.188	& 0.211	& 0.208	& 0.181	& 0.144	& 0.178	& 0.169 \\ 
MLP    & 0.184	& 0.215	& 0.182 & 0.191	& 0.216	& 0.191	& 0.203	& 0.214	& 0.190 & 0.191	& 0.184	& 0.178\\ \midrule
\multicolumn{13}{l}{\textbf{Cross-city Transfer baselines}}    \\
MetaST  & 0.232	& 0.231	& 0.209	& 0.232	& 0.225	& 0.203	& 0.204	& 0.226	& 0.202	& 0.219	& 0.204	& 0.197 \\
MetaStore  & 0.236	& 0.234	& 0.216	& 0.235	& 0.222	& 0.214	& 0.236	& 0.233	& 0.212	& 0.225	& 0.212	& 0.207 \\ \midrule
\multicolumn{13}{l}{\textbf{Cross-city Federated Transfer baselines}}    \\
CNN+FedAvg  & 0.225	& 0.224	& 0.194	& 0.209	& 0.212	& 0.195	& 0.212	& 0.222	& 0.197	& 0.192	& 0.206	& 0.180 \\
MLP+FedAvg  & 0.231	& 0.226	& 0.203	& 0.226	& 0.215	& 0.203	& 0.213	& 0.227	& 0.203	& 0.197	& 0.201	& 0.191 \\ \midrule
\multicolumn{13}{l}{\textbf{Ours}}    \\
CcFTL(CNN)  & 0.243	& 0.244	& 0.222	& \textbf{0.256}	& 0.247	& 0.233	& 0.262	& 0.236	& 0.221	& 0.253	& \textbf{0.225}	& 0.204 \\
CcFTL(MLP) & \textbf{0.269}	& \textbf{0.250}	& \textbf{0.248}	& 0.248	& \textbf{0.254}	& \textbf{0.237}	& \textbf{0.421}	& \textbf{0.241}	& \textbf{0.231}	& \textbf{0.294}	& 0.221	& \textbf{0.213} \\ \bottomrule
\end{tabular}}
\label{tab2}
\end{table*}

\subsubsection{Implementations}
We implement XGBoost with Sklearn library, and CNN, MLP, CNN+FedAvg, MLP+FedAvg, and CcFTL using the machine learning library Pytorch 1.2 in Python 3.6. 
Besides that, we implement MetaST and MetaStore with the machine learning library Tensorflow 1.13. 

In the implementation of our framework CcFTL, we set the numbers of the global rounds and local training epochs in the training phase are 200 and 1, respectively. 
For model optimization, we take SGD as our optimization algorithm. 
For the hyperparameters of the optimizer, we set the learning rate as 0.01, and the batch size as 128. 
For the DARKL module, we set the dimensions of the two fully connected layers in the \textit{Feature Extractor} to be 256 and 128 respectively, the dimensions of the two fully connected layers and one regression layer in the \textit{Data Regressor} to be 64, 32 and 1, respectively, and the dimensions of one fully connected layer and one classification layer in the \textit{Domain Classifier} to be 64 and the number of domains (source cities), respectively. 
For the UTP module, we set the dimensions of the three fully connected layers and one classification layer to be 256, 64, 32, and 5 respectively. 
For the loss function, we set $\lambda$ as 0.6. 
Finally, we fine-tune 50 epochs in the target city. 

In the implementations of baselines, we maintain the number of training epochs on multiple source cities as 200, the learning rate as 0.01, and the batch size as 128. 
For other implementation details in the baselines, as follows: 
1) For XGBoost, we set the number of estimators to 100, the maximum depth to 6, the gamma parameter to 0, and the reg\_lambda parameter to 1. 
2) For CNN, it consists of 3 convolutional layers, using 32 convolution kernels of 2$\times$, 3$\times$, and 4$\times$ respectively. The dropout ratio is set as 0.1 at the model training stage. 
3) For MLP, we set the dimensions of the three fully connected layers and one classification layer to be 128, 64, and 32 respectively, and use the $\textrm{ReLU}$ activation function. 
4) For MetaST, we set the meta-learning rate as $10^{-2}$. Besides, we use K-means to cluster the spatio-temporal patterns of all regions into 4 groups and set the size of pattern representation in memory as 64.
5) For MetaStore, we take the same base model as MetaST and also set the meta-learning rate to $10^{-2}$. We learn each city embedding vector with unsupervised learning approach and set its size to 128. 
6) For CNN+FedAvg and MLP+FedAvg, we use CNN and MLP as the base model respectively and also maintain the numbers of the global rounds and local training epochs as CcFTL. 
7) For CcFTL(CNN), in the DARKL module, the \textit{Feature Extractor} consists of three convolutional layers and one fully connected layer, the \textit{Data Regressor} consists of two convolutional layers and one regression layer, and the \textit{Domain Classifier} consists of one convolutional layer and one classification layer. 
In the UTP module, we use 3 convolutional layers and one classification layer. 

\subsection{Results}
\subsubsection{Overall Performance} 
We compare our CcFTL with the baseline models in terms of \textit{Precision}, \textit{Recall}, and \textit{F1 score} in the task of predicting consumption power. 
The results are shown in Table~\ref{tab2}. 
Note that (Ny, Xc) $\to$ Ly is an abbreviation for (Nanyang, Xuchang) $\to$ Luoyang, and the same is true for others. 
We have the following observations according to the analysis of the results in the table: 
\begin{itemize}[leftmargin=*]
\item Target-only baselines are trained based on limited data in four target cities. 
XGBoost struggles to handle complex urban computing tasks under data insufficiency conditions. 
The performance of CNN is slightly weaker than that of MLP, which is more suitable for two-dimensional or multi-dimensional data features. 
MLP achieves the best performance in Target-only baselines, which can learn non-linear relationships better. 
While our proposed method CcFTL consistently outperforms Target-only baselines in terms of three evaluation metrics. 
This indicates the CcFTL can learn general relational knowledge across cities and apply it in the target city to address data insufficiency problems effectively. 
\item MetaST and MetaStore are competitive and state-of-the-art cross-city transfer methods, which are trained in a centralized manner. 
Our method CcFTL outperforms them. 
Experimental results demonstrate that existing cross-city transfer methods are difficult to adapt to complex data-insufficient scenarios (data missing and label scarcity), and do not consider the issue of urban data privacy. 
Our approach not only protects urban data security but also effectively addresses the challenge of insufficient data in target cities. 
\item Both CNN+FedAVG and MLP+FedAVG train a global model from multiple source cities under federated learning settings and transfer it to the target city for fine-tuning. 
The performance of MLP+FedAVG is slightly better than that of CNN+FedAVG, which we believe is because MLP has a stronger ability to capture nonlinear relationships than CNN. 
Experimental results show that CcFTL is much better than federated transfer baselines. 
From these results, we can indicate that simply applying federated learning to cross-city scenarios cannot effectively solve urban computing problems, reflecting the innovativeness and importance of relational knowledge learning in our paper. 
\item CcFTL(CNN) is a variant of our framework that uses CNN as the structure of the base model instead of MLP. 
CcFTL(MLP) and CcFTL(CNN) together achieve the best performance, which not only shows the effectiveness of our framework, but also the generalizability of the framework. 
Our framework can be applied to various base models to solve different urban computing tasks under privacy protection, with good compatibility. 
\begin{figure}[t]
\centering
\includegraphics[width=1.12\linewidth, trim=65 0 0 0, clip]{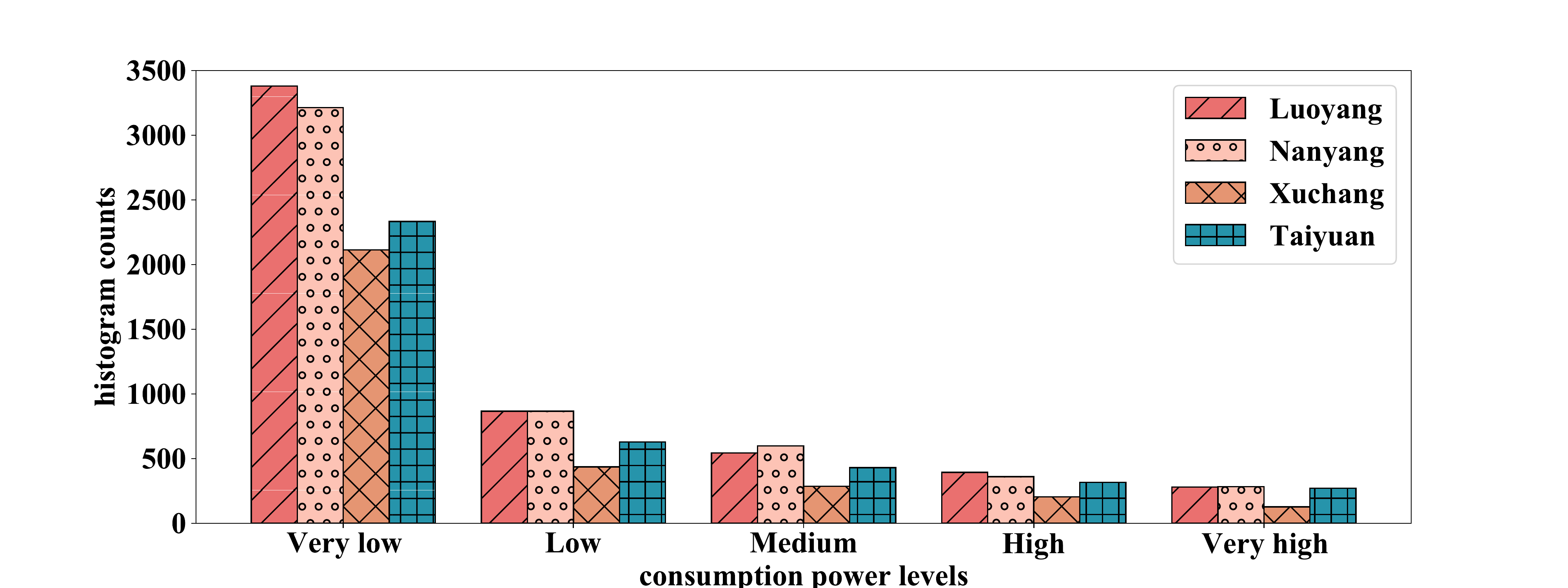}
\caption{Long-tailed distribution of regions with different consumption power levels in cities.}
\label{Figure 8}
\end{figure}
\item Our experimental results do not seem impressive, due to the long-tailed distribution of the labels of the consumption power in the city (as shown in Figure~\ref{Figure 8}). 
There are many regions with low consumption power levels and few regions with high consumption power levels, which leads to imbalanced training samples. 
And we compute the macro metrics in terms of \textit{precision}, \textit{recall}, and \textit{F1 score}. 
How to improve prediction performance based on long-tailed data in urban computing, which can be a future research. 
\item Our approach CcFTL attains the best performance across all of the evaluation metrics with different city pairs. 
Specifically, in the target city of Luoyang, \textit{Precision}, \textit{Recall}, and \textit{F1 Score} of CcFTL are relatively improved outperforming the best baseline by 13.9\%, 6.8\%, and 14.8\%. 
In the target city of Nanyang, \textit{Precision}, \textit{Recall}, and \textit{F1 Score} of CcFTL are relatively improved outperforming the best baseline by 8.9\%, 12.8\%, and 10.7\%. 
In the target city of Xuchang, \textit{Precision}, \textit{Recall}, and \textit{F1 Score} of CcFTL are relatively improved outperforming the best baseline by 78.3\%, 3.4\%, and 8.9\%. 
In the target city of Taiyuan, \textit{Precision}, \textit{Recall}, and \textit{F1 Score} of CcFTL are relatively improved outperforming the best baseline by 30.6\%, 6.1\%, and 2.8\%. 
These experimental results prove the effectiveness of our CcFTL in cross-city federated transfer learning scenarios. 
\end{itemize}

If there is no special description, CcFTL appearing in the following refers to CcFTL(MLP). 
\subsubsection{Ablation Study} 
\begin{table}[t]
\caption{Ablation studies of our method CcFTL on on two city pairs. We observed similar results in other city pairs.}
\resizebox{\linewidth}{!}{
\begin{tabular}{lcccccc}
\toprule
\multicolumn{1}{c}{}                                & \multicolumn{3}{c}{\textbf{(Ty, Ly) $\to$ Ny}} & \multicolumn{3}{c}{\textbf{(Xc, Ny) $\to$ Ty}}  \\ \cmidrule(r){2-4} \cmidrule(r){5-7} 
\multicolumn{1}{c}{}                                & P   & R    & F1  & P   & R    & F1    \\ \midrule
w/o DARKL   & 0.226	& 0.215	& 0.203	& 0.197	& 0.201	& 0.191 \\ 
w/o UTP    & 0.231	& 0.235	& 0.218	& 0.246	& 0.214	& 0.200 \\
w/o Fine-tuning   & 0.236	& 0.238	& 0.229	& 0.253	& 0.215	& 0.204 \\
\textbf{CcFTL(MLP)} & \textbf{0.248}	& \textbf{0.254}	& \textbf{0.237}	& \textbf{0.294}	& \textbf{0.221}	& \textbf{0.213}	\\ \bottomrule
\end{tabular}}
\label{tab3}
\end{table}
We further investigate that the Domain Adaptive Relational-Knowledge Learning (DARKL) module, Urban Task-specific Predicting (UTP) module, and fine-tuning strategy are essential parts of our framework CcFTL. 
As shown in Table~\ref{tab3}, we can observe that: (1) Without the DARKL module, the F1 score drops from 23.7\% to 20.3\% in Nanyang, 21.3\% to 19.1\% in Taiyuan. 
The experimental results show that the DARKL module contributes the largest increase, which also that proves that the DARKL module can greatly alleviate the challenge of data missing in the target city. 
(2) We only learn relational knowledge from multiple source cities without learning urban task-specific parameters, and transfer relational knowledge directly to the target city, to demonstrate the gain brought by the UTP module. 
The UTP module contributes to 8.7\% and 6.5\% relatively increase in terms of F1 score on Nanyang and Taiyuan respectively, as it can bring a better initialization parameter for urban computing task for the target city. 
(3) Fine-tuning strategy improves the relative 3.4\% and 4.4\% performance in terms of F1 score on Nanyang and Taiyuan respectively, because it can obtain a personalized adaptation to the target city. 

\subsubsection{Security vs Performance} 
\begin{table}[t]
\caption{F1 score performance of baselines trained in a centralized manner and our method on four city pairs. They all use MLP as the base model.}
\resizebox{\linewidth}{!}{
\begin{tabular}{lcccc}
\toprule
\multicolumn{1}{c}{} & \textbf{(Ny, Xc)} & \textbf{(Ty, Ly)} & \textbf{(Ly, Ty)} & \textbf{(Xc, Ny)} \\
\multicolumn{1}{c}{} & \textbf{$\rightarrow$ Ly}  & \textbf{$\rightarrow$ Ny}  & \textbf{$\rightarrow$ Xc}  & \textbf{$\rightarrow$ Ty}  \\ \midrule
MetaST               & 0.209   & 0.203   & 0.202   & 0.197    \\
MetaStore            & 0.216   & 0.214   & 0.212   & 0.207   \\
CcFTL(MLP)           & 0.248   & 0.237   & 0.231   & 0.213   \\
CcFTL(centralized)           & \textbf{0.257}   & \textbf{0.240}   & \textbf{0.237}   & \textbf{0.233}   \\ \bottomrule
\end{tabular}}
\label{tab7}
\end{table}
CcFTL(centralized) is a model directly trained with centralized data, which architecture is exactly the same as CcFTL. We see its performance as the upper bound of the performance of CcFTL. 
That is to say, CcFTL(centralized) does not take into account the privacy protection of the urban data. 
In addition, MetaST and MetaStore are also trained in a centralized manner, and these methods all use MLP as the base model. 
We compare the performance of these methods with ours on four city pairs and present the F1 score metric in Table~\ref{tab7}.
CcFTL(centralized) consistently outperforms MetaST and MetaStore, which demonstrates the effectiveness of relational knowledge learning and cross-city transfer in our framework. 
Compared with CcFTL(centralized), our model CcFTL has an absolute decrease of 0.9\%, 0.3\%, 0.6\% and 2\% in the F1 score metric of the four city pairs, respectively. 
That demonstrates that our framework enables secure cross-city data sharing and utilization with acceptable performance loss.  

\subsubsection{Importance of the Domain Classifier}
\begin{figure}[t]
\centering
\includegraphics[width=\linewidth, trim=0 50 0 00, clip]{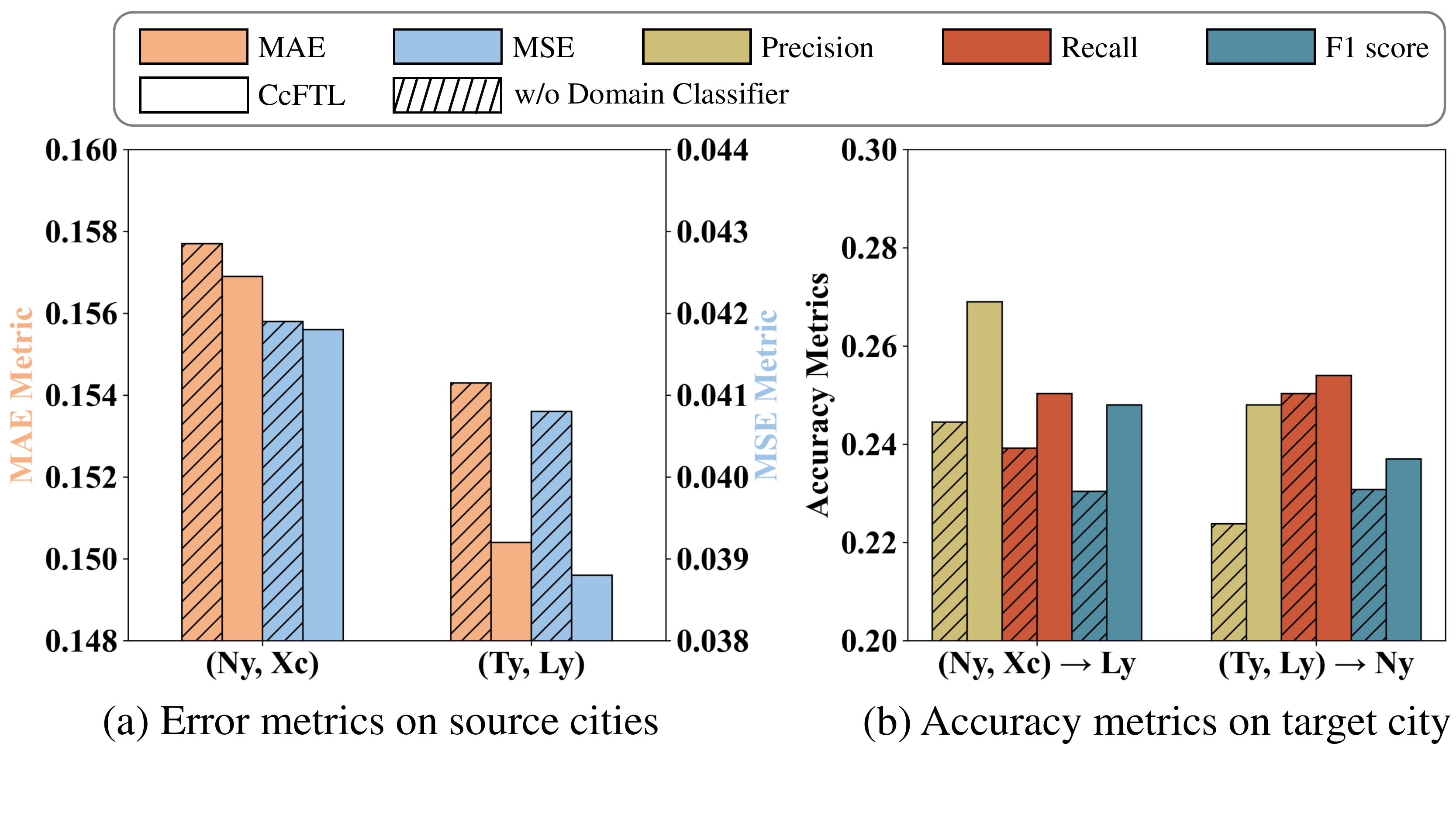}
\caption{Comparison of error metrics and accuracy metrics for the DARKL module with and without domain classifier.}
\label{Figure 4}
\end{figure}
To verify that the \textit{Domain Classifier} can drive the \textit{Feature Extractor} to learn feature representations that generalize to domains, we compare the performance of CcFTL with and without \textit{Domain Classifier}. 
On the one hand, as shown in Figure~\ref{Figure 4} (a), orange bars and the left axis represent MAE metrics, blue bars and the right axis represent MSE metrics, and bars with stripes represent performance without the \textit{Domain Classifier}. 
Adding domain classifier can effectively reduce the error metrics \textit{MAE} and \textit{MSE} when learning relational knowledge based on multi-source data of city pairs (Nanyang, Xuchang) and (Taiyuan, Luoyang). 
Since the data is normalized, small differences in error metrics can be huge in practicality. 
On the other hand, as shown in Figure~\ref{Figure 4} (b), the relational knowledge learned with the \textit{domain classifier}, and transferred to the target city can greatly improve the accuracy of the region profiling prediction task. 
Experimental results show that our method CcFTL can effectively alleviate domain differences while learning common relational knowledge across domains and benefit the target city.

\subsubsection{Scalability Analysis} 
\newcommand{\tabincell}[2]{\begin{tabular}{@{}#1@{}}#2\end{tabular}}
\begin{table}[t]
\caption{Comparison of error metrics and average training epoch time during training of the different number of clients.}
\resizebox{1\linewidth}{!}{
\begin{tabular}{cccc}
\toprule
 & MAE  & MSE  & \tabincell{c}{Average Training\\Epoch Time (s)} \\ \midrule
(Ly, Ny) & 0.1559 & 0.0417 & 23.69 \\
(Ly, Ny, Xc) & 0.1542 & 0.0406 & 30.65 \\
(Ly, Ny, Xc, Ty) & 0.1627 & 0.0452 & 40.85 \\ \bottomrule
\end{tabular}
}
\label{tab4}
\end{table}
Our CcFTL framework is theoretically scalable, allowing more cities to participate as subjects. 
However, the application of the cross-city federated transfer learning framework in a certain number of cities with such different distributions is bound to bring performance and efficiency disadvantages. 
Thus, we compare the impact of different numbers of clients on the Domain Adaptive Relational-Knowledge Learning module in terms of error metrics and training time in Table~\ref{tab4}. 
As the number of source cities participating in the training stage increases, our framework CcFTL can effectively alleviate the differences among cities. 
However, when the number of source cities increases to a certain extent, the data distributions are so different that the error metric inevitably rises, and the time cost increases faster. 
Thus, although our framework is scalable and can be deployed to multiple source cities, the number of participants in the training stage is still an important factor. 
That is one of the reasons why we adopt the experimental setup of learning knowledge from two source cities and transferring it to the target city in the main experiments. 

\subsubsection{The impact of Label Scarcity}
\begin{figure}[t]
\centering
\includegraphics[width=\linewidth, trim=0 100 0 00, clip]{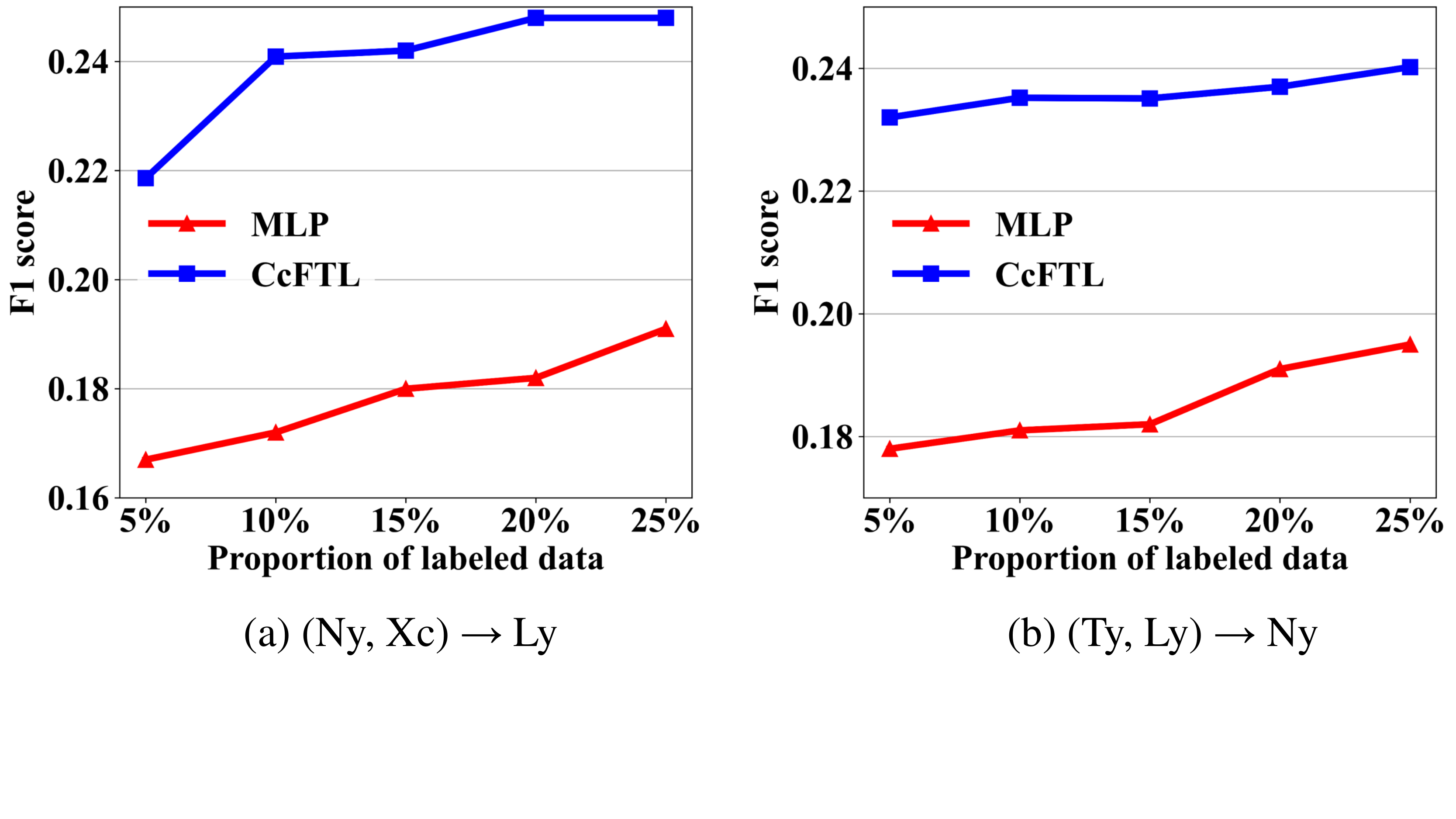}
\caption{Comparison of MLP and CcFTL performance on different city pairs, as the proportion of available labeled data increases.}
\label{Figure 5}
\end{figure}
We further explore the effectiveness of our model CcFTL under varying proportions of label scarcity in the target city. 
Specifically, we compare the performance of MLP and CcFTL under the following settings: for a target city, we randomly select 5\% to 25\% of the labeled data as the training sets and the rest as the test sets, step is 5\%. 
As shown in Figure~\ref{Figure 5}, the blue line represents the performance change in terms of the F1 score metric of CcFTL, while the red line represents the MLP. 
As the proportion of labeled data in the target city increases, the performance of CcFTL and MLP also improves. 
However, we observe that the improvement brought by CcFTL is much larger in the target city Nanyang than in Luoyang. 
We argue that this is because the development level of Luoyang is much higher than that of Nanyang, so Nanyang is more sensitive to multi-source data than Luoyang in the urban computing task of predicting consumption power.

\subsubsection{$\lambda$ Parameter Tuning}
\begin{figure}[t]
\centering
\includegraphics[width=\linewidth, trim=0 120 0 00, clip]{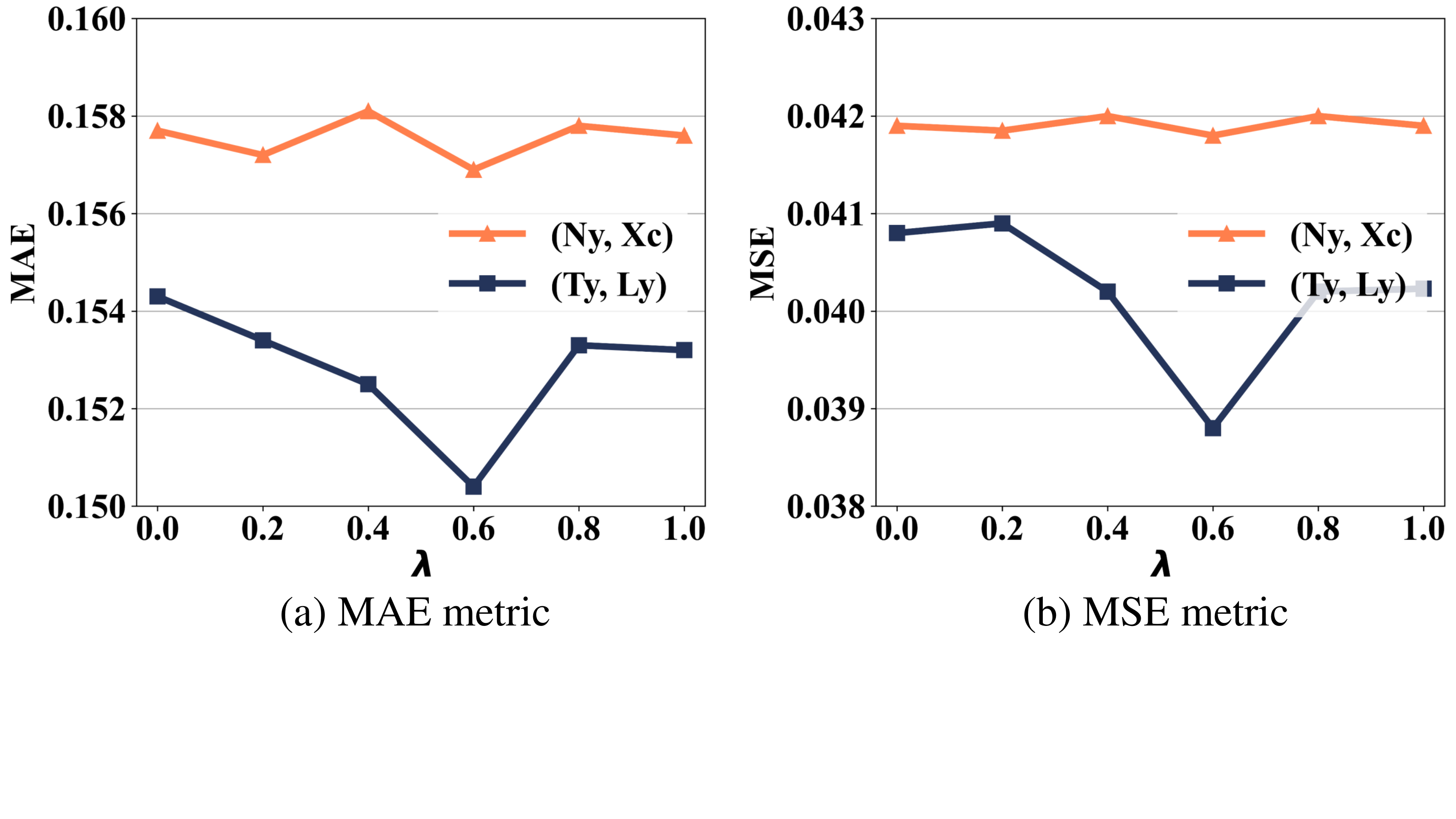}
\caption{The impact of hyperparameter $\lambda$ on DARKL module training.}
\label{Figure 6}
\end{figure}
$\lambda$ is a hyperparameter in the loss function of the Domain Adaptive Relational-Knowledge Learning module, which controls the weight of the \textit{domain classifier} during training. 
To show the impact of the parameter $\lambda$, we evaluate our model under different $\lambda$ in the range of [0, 1], step is 0.2. 
We show the results in terms of MAE and MSE for different $\lambda$ on city pairs during the federated training stage in Figure~\ref{Figure 6}. 
The orange line represents the (Nanyang, Xuchang) city pair, and the dark blue line represents the (Taiyuan, Luoyang) city pair. 
We can observe that as $\lambda$ increases, CcFTL can effectively alleviate the domain difference and reduce the error metrics. 
However, when $\lambda$ is too large, the domain classifier acts as a negative learner. 
Therefore, selecting an appropriate parameter $\lambda$ is crucial for the performance of the model. 
In this work, we set $\lambda$ to 0.6. 

\subsection{Case Study}
\begin{figure}[t]
\centering
\includegraphics[width=\linewidth]{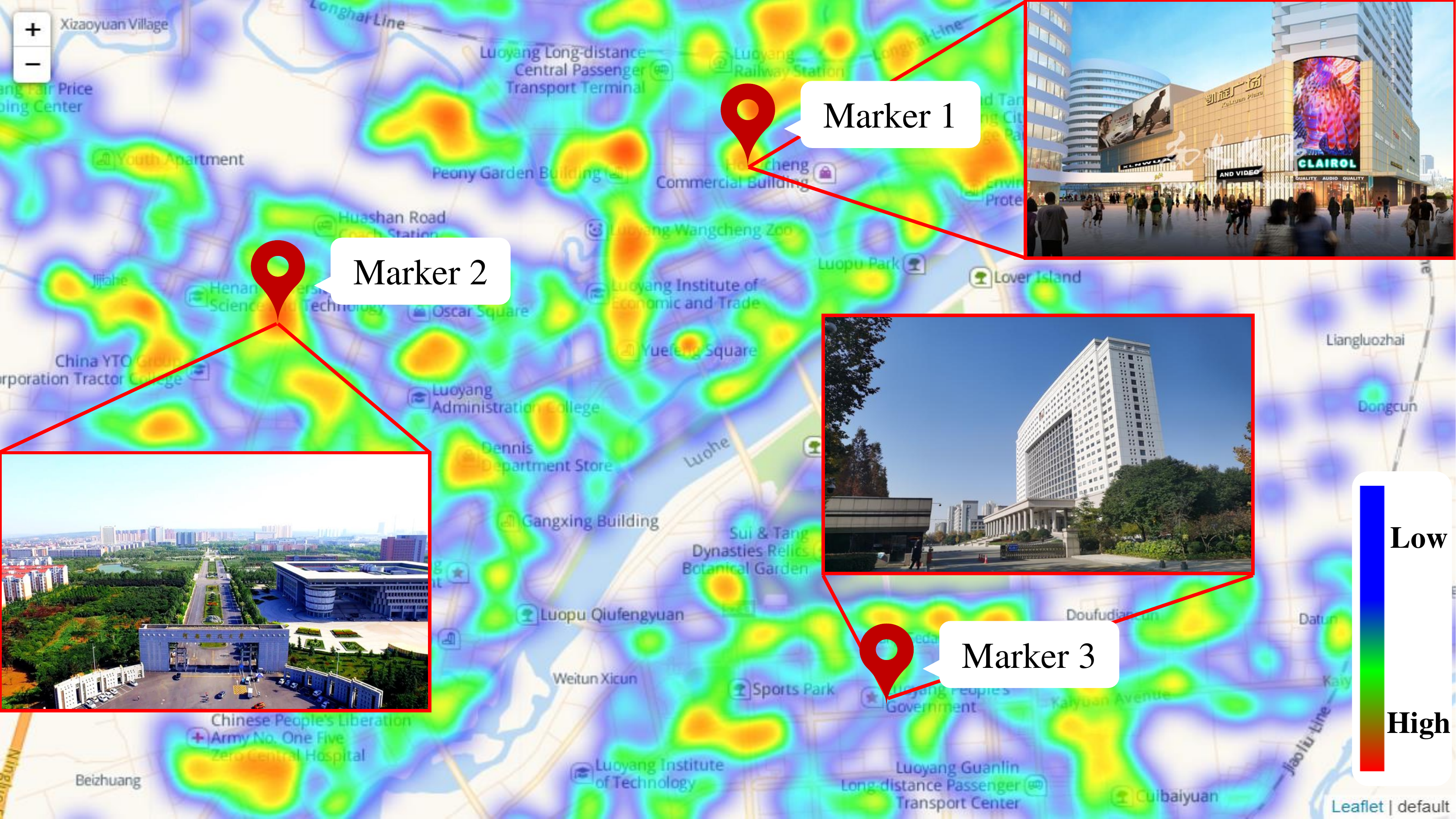}
\caption{A Case Study in Luoyang.}
\label{Figure 7}
\end{figure}

We further give a case study of predicting the consumption power of Luoyang. 
We plot the prediction results for the central area of Luoyang as a heat map and show it in Figure~\ref{Figure 7}.
Red represents a higher consumption power level while blue represents a lower level. 

We take three markers on the heat map as examples for analysis, which can provide insights for smart city applications such as urban planning and business district location selection. 
Marker 1 is the Hongcheng Commercial Building that attracts a large number of consumption populations. 
It drives a variety of businesses to settle in and forms a business district. 
Thus, the consumption power of the surrounding area is very high, and there are many other educational institutions around there, as well as public institutions such as hospitals. 
Such geographical location and environment attract a large number of residential populations and drive the consumption power of the whole area. 
And marker 3 is Luoyang Peoples Government, and there are no entertainment venues nearby, so the relative consumption power is low. 
For the city of Luoyang, business districts can be planned near Marker 1 and Marker 2 to promote economic development. 
These also demonstrate that our model CcFTL can work in the real world to facilitate the construction of smart cities.

\section{Related Work}
In this section, we introduce the related work in three aspects: urban computing, federated learning, and transfer learning. 

\subsection{Urban Computing}
Urban computing~\cite{zheng2014urban} can help city manager (i.e. government) improve people’s life quality, protect the environment, and promote the efficiency of city operation. 
Urban computing continuously acquires, integrates, and analyzes multi-source data in the city, to solve the challenges faced by the city, such as air quality prediction~\cite{han2021joint} and traffic flow prediction~\cite{pan2020spatio}. 
Meanwhile, urban region profiling is also one of the spatio-temporal prediction applications.  
It describes the state of the city, such as business popularity and traffic convenience. 
Specifically, the traffic convenience of the region can help the government to design better transportation scheduling~\cite{fang2021spatial}, and the consumption power and business popularity are also very helpful for business district planning~\cite{xu2020ar2net}.
Therefore, achieving accurate urban region profiling can identify problems in the construction process of smart city and directions for improvement. 

\subsection{Federated Learning}
Federated learning is a machine learning setting where multiple clients train a model in collaboration with decentralized training data, which was firstly proposed by Google~\cite{konevcny2016federated}. 
Many researchers 
have successfully applied federated learning to medical~\cite{yang2021flop}, recommendation~\cite{wang2021poi, guo2021prefer}, traffic prediction~\cite{DBLP:conf/kdd/MengRL21}, and other fields. 
Federated learning can resolve the data islanding problems by privacy-preserving model training in the network. 

According to~\cite{yang2019federated}, federated learning can mainly be classified into three types: 
1) horizontal federated learning, where organizations share partial features.
2) vertical federated learning, where organizations share partial samples.
and 3) federated transfer learning, where neither samples nor features have much in common.

CcFTL belongs to federated transfer learning category. 
It is the first of this kind tailored for cross-city federated transfer framework. 

\subsection{Transfer Learning}
Transfer learning aims to extract the knowledge from one or more source tasks and applies the knowledge to a target task. 
In the setting of transfer learning, the domains are often different but related, which makes knowledge transfer possible. 
The key idea is to reduce the distribution divergence between different domains. 

To this end, there are mainly four kinds of approaches: 
1) instance-based transfer~\cite{wang2019minimax, huan2021learning}, which assumes that certain parts of the data in the source domain can be reused for learning in the target domain by re-weighting; 
2) feature-representation transfer~\cite{yin2019feature, zhou2019deep, lu2020cross}. 
The intuitive idea behind this case is to learn a “good” feature representation for the target domain; 
3) parameter transfer~\cite{rozantsev2018residual, houlsby2019parameter, yuan2020parameter}, which assumes that the source tasks and the target tasks share some parameters or prior distributions of the hyper-parameters of the models. 
And 4) relational-knowledge transfer~\cite{ge2020look}, which assumes that some relationships among the data in the source and target domains are similar.

CcFTL is mainly related to the relational-knowledge transfer. 
On the one hand, although the previous method FLORAL~\cite{wei2016transfer} is based on the instance and parameter transfer to address the data insufficiency in the target city, FLORAL is not designed for deep learning, and our framework CcFTL is scalable and extensible. 
On the other hand, most methods of transfer learning across cities assume the availability of training data, which is not realistic. 
CcFTL makes it possible to do deep transfer learning in the federated learning framework without accessing the raw user data.

\section{Conclusion and Future Work}
In this paper, we propose CcFTL, the first cross-city federated transfer learning framework. 
CcFTL aggregates the multi-source data from multiple cities without compromising privacy security and alleviates data insufficiency challenges (data missing and label scarcity) in the target city through relational knowledge transfer and parameters transfer. 
Experiments on consumption power prediction in urban region profiling have demonstrated the 
effectiveness of the framework. We also present a case study to show that our model can provide insights for smart city construction.  

CcFTL opens a new door for future research in smart cities.
As mentioned in the analysis of the experimental results, the long-tailed distribution of urban data in the cross-trial federation transfer 
scenario can be one of the future research directions. 

\bibliographystyle{IEEEtran}
\bibliography{IEEEabrv, myrefs}

\end{document}